\documentclass[3p]{elsarticle}

\usepackage{amsmath,amssymb,amsfonts}
\usepackage[ruled,linesnumbered]{algorithm2e}
\usepackage{graphicx}
\usepackage{subfigure}
\usepackage{multirow}
\usepackage{booktabs}
\usepackage{tabularx}
\usepackage{textcomp}
\usepackage{xcolor}
\usepackage{float}
\usepackage[colorlinks,
            linkcolor=blue,
            anchorcolor=blue,
            citecolor=blue]{hyperref}
\usepackage{diagbox}
\usepackage{epstopdf}
\usepackage{amsmath}
\usepackage{amsfonts}
\journal{ }

\begin{document}

\begin{frontmatter}



\title{Scalable Multiple Patterning Layout Decomposition Implemented by a Distribution Evolutionary Algorithm \tnoteref{grant}}
\tnotetext[grant]{This work was supported in part by the National Key R\& D Program of China (2021ZD0114600).}

\author[label1]{Yu Chen}
\ead{ychen@whut.edu.cn}

\author[label1]{Yongjian Xu}
\ead{xyjcbhl88888@163.com}

\author[label2]{Ning Xu\corref{cor1}}
\ead{Xuning@whut.edu.cn}
%\affiliation[label1]{organization={School of Science, Wuhan University of Technology},%Department and Organization
%            city={Wuhan},
%            postcode={430070},
%            state={Hubei},
%            country={China}}
%\affiliation[label2]{organization={School of Information Egineering, Wuhan University of Technology},%Department and Organization
%            city={Wuhan},
%            postcode={430070},
%            state={Hubei},
%            country={China}}
\cortext[cor1]{Corresponding author.}
\begin{abstract}
As the feature size of semiconductor technology shrinks to 10 nm and beyond, the multiple patterning lithography (MPL) attracts more attention from the industry. In this paper, we model the layout decomposition of MPL as a generalized graph coloring problem, which is addressed by a distribution evolutionary algorithm based on a population of probabilistic model (DEA-PPM). DEA-PPM can strike a balance between decomposition results and running time, being scalable for varied settings of mask number and lithography resolution. Due to its robustness of decomposition results, this could be an alternative technique for multiple patterning layout decomposition in next-generation technology nodes.
\end{abstract}

%\includegraphics{MPL_FLOW}

%\item Addressing the multiple patterning layout decomposition problem by solving the graph coloring model with efficient metaheuristics.
%\item Proposing an efficient algorithm scalable to a variety of MPLD problems.


\end{frontmatter}


%\section{}
%\label{}
\section{Introduction}
Multiple patterning lithography (MPL) improves the resolution limit of existing lithography technologies, which could then meet the challenge of further decrease of feature size for the next-generation technology nodes~\cite{Ma2017MethodologiesFL}.
Given a layout specified by features in polygonal shapes, a layout graph (LG) is an undirected graph whose nodes are the given layout's features and where an edge exists if and only if two polygonal shapes are beyond the lithography resolution limit~\cite{kahng2008layout}. Multiple patterning layout decomposition (MPLD) partitions a layout into several sections (masks), which is then modelled as a graph coloring problem (GCP).

The GCP can be generally formulated as an integer optimization problem, and a variety of integer linear programming (ILP) methods have been developed to address the MPLD problem~\cite{Yu2011LayoutDF,Yu2014TriplePL,Lin2017ColorBA,Li2017DiscreteRM}. Since the GCP is NP-complete, the ILP methods cannot decompose large-scale layouts efficiently, and  it was relaxed to a semidefinite programming (SDP)~\cite{Yu2013AHT} problem or a linear programming (LP) problem~\cite{Lin2017TriplequadruplePL}. Although the SDP and LP methods leads to scalability of MPLD to large-scale cases, additional repair processes are needed to get the feasible solutions of GCP, and the relaxation would mathematically lead to solutions different from those of the original MPLD problem. Accordinly, the MPLD problem was also addressed by heuristics~\cite{Fang2012ANL,Kuang2013AnEL,Zhang2015LayoutDW,Ke2015AntCA,Chen2019PrintabilityEW}, and some layout decomposition works based on hybrid lithography have been investigated~\cite{Yang2015LayoutDC,Li2021DiscreteRM}.

According to the number of colors (masks), MPLD problems can be categorized as the double pattering layout decomposition (DPLD) problem~\cite{Yuan2010,Yang2010,Kahng2010,Hsu2011,Zhao2014}, the triple pattering layout decomposition (TPLD) problem~\cite{yu2014LC,Fang2014TPL,Yu2015TPL,YU2015LD,Zhang2015LD,Ke2016NC,Kohira2016} and the quadruple pattering layout decomposition (QPLD) problem~\cite{Yu2014LayoutDF}, etc. Because the uncolorable modules for varied settings of color number contain different connection topologies,
most of the algorithms for MPLD problems are not scalable for varied settings of color number. Yu and Pan~\cite{Yu2014LayoutDF} proposed an SDP model for QPLD, where extra color assignment strategies are introduced, and  the proposed algorithm could be applicable to a $k$ patterning layout decomposition problem with $k\ge 4$.
Lin \emph{et al,}~\cite{Lin2017TQP} proposed a new and effective algorithm for TPLD and QPLD. By utilizing the features in the polyhedron space of the feasible set, the ILP formulation is approximated by the LP relaxation. Jiang and Chang~\cite{Hui-Ru2017MPL} modelled the MPLD problem as an exact cover problem to construct a fast and exact MPLD framework based on augmented dancing links, which can consider the basic and complex coloring rules simultaneously, can maintain density balancing, and can handle quadruple patterning and beyond.

Recently, Li \emph{et al.}~\cite{Li2019OpenMPLAO,9279240} proposed an open source layout decomposer named as the \emph{OpenMPL}, where a general framework of MPLD is developed for varied settings of mask number and several alternative optimization schemes including the ILP and the SDP. Although \emph{OpenMPL} incorporates several simplification strategies of the layout graph, the ILP algorithm still suffers from high computational complexity and the solution of SDP is of low quality as usual. In order to develop a scalable MPLD framework that strikes a good balance between the running time and the solution quality, we propose to address it via a distribution evolutionary algorithm based on a population of probability distribution (DEA-PPM)~\cite{xu2022distribution}, which is based on a novel probability model updated by an orthogonal transformation and a gradual renewal strategy of the distribution model. Meanwhile, the solution space is exploited by deploying a related solution population refined by a tabu search (TS). As a result, the DEA-PPM relying on small populations is expected to achieve robust results on varied settings of the MPLD problem.

%\subsection{DPL}
%
%
%
%
%
%Yuan \emph{et al.}~\cite{Yuan2010} proposed a grid model to decompose the layout into grids, by which an integer linear programming model was developed to address the double-patterning layout decomposition problem.
%
%Yang \emph{et al.}~\cite{Yang2010} developed a multi-objective min-cut based decomposition framework for stitch minimization, balanced
%density, and overlay compensation, simultaneously.
%
%Kahng \emph{et al.}~\cite{Kahng2010} address this problem using two layout decomposition approaches based on a conflict graph. First, node splitting is performed at all feasible dividing points. Then, one approach detects conflict cycles in the graph which are unresolvable for DPL coloring, and determines the coloring solution for the remaining nodes using integer linear programming (ILP). The other approach, based on a different ILP problem formulation, deletes some edges in the graph to make it two-colorable, then finds the coloring solution in the new graph.
%
%Hsu \emph{et al.}~\cite{Hsu2011} proposed to simultaneously perform the
%layout migration and the decomposition problem (Layout compliance).
%
%Zhao \emph{et al.}~\cite{Zhao2014} proposed to address the double patterning decomposition of layout in a parallel way to accelerate the decomposition process and to reduce the peak memory consumption.

The remainder of this paper is organized as follows.
Section \ref{SecMPL} briefly introduces the framework of \emph{OpenMPL}. In Section \ref{SecDEA}, the details of DEA-PPM are presented,  and  Section \ref{SecRes} verifies the competitiveness of DEA-PPM by experimental results. Finally, we conclude the work in Section \ref{SecCon}.

\section{Multiple Patterning Layout Decomposition based on the OpenMPL}\label{SecMPL}
\subsection{The Multiple Patterning Layout Decomposition Problem}

To address the MPLD problem,the layout is represented by a decomposition graph (DG) consisting of a set of nodes $V$ and two sets of edges, $CE$ and $SE$, which contain the conflicting edges and stitch edges, respectively \cite{yuan2010wisdom}. Accordingly, the goal of MPLD is to assign nodes of $V$ with $k$ colors, trying to achieve a nice trade-off between the numbers of conflict and stitch, respectively defined by

\begin{equation}
c_{uv} =\begin{cases}
  & \text{1} \quad \mbox{if } (u,v)\in CE,  \\
  & \text{0} \quad \mbox{otherwise,}
\end{cases}
\end{equation}
and
\begin{equation}
s_{uv} =\begin{cases}
  & \text{1} \quad \mbox{if } (u,v)\in SE,  \\
  & \text{0} \quad \mbox{otherwise.}
\end{cases}
\end{equation}
Then, minimization of conflicts and stitches can be formulated as
\begin{equation}\label{model}
min\quad f(\mathbf{x})=\sum_{\left( u,v \right)\in CE}c_{uv} +\alpha\times\sum_{\left( u,v\right)\in SE}s_{uv},
\end{equation}
where $\alpha$ is the weight parameter, set as $0.1$ in this paper~\cite{9279240}. For cases $k=2,3,4$, problem (\ref{model})  corresponds to the DPLD problem, the TPLD problem and the QPLD problem, respectively.

\subsection{Framework of the OpenMPL}
As presented in \emph{OpenMPL}~\cite{Li2019OpenMPLAO,9279240}, the work flow of  MPLD starts with generation of the decompositon graph $G$. Then, size of graph $G$ is reduced by the graph simplification operations, and candidate stitches are inserted to get a modified conflicting graph $G'$. Before addressing the layout decomposition, $G'$ is further simplified to a reduced graph $G''$. Establishing an optimization model for $G''$, one can employ a selected optimization algorithm to get the colored graph $G''$. At last, the decomposition result of layout can be obtained by recovering the coloring result of $G''$ to that of the original conflicting graph $G$. Details of the framework of \emph{OpenMPL} are referred to \cite{Li2019OpenMPLAO,9279240}.

\begin{figure}[htb]
\centering
{
%\subfigure[MND algorithm, $Re-ssd=47.70\%$, $zz=32.6\%$, $MFN=0$ Box plot of running time for \emph{DSJC125.5}]{
\includegraphics[width=4.5cm]{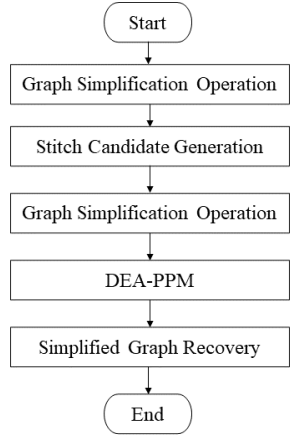}
}
\caption{The workflow of layout decomposition based on the \emph{OpenMPL}.}
\label{MPLflow}
\end{figure}

\emph{OpenMPL} incorporates an ILP approach, an SDP approach and a flexible exact cover-based approach as three alternative routines of MPLD. To improve the efficiency of layout decomposition, we proposed to address it by a distribution evolutionary algorithm based on a population of probability models (DEA-PPM)~\cite{xu2022distribution}, and get the work flow of layout decomposition illustrated in Fig. \ref{MPLflow}.

\section{A Distribution Evolutionary Algorithm Based
on a Population of Probability Model}\label{SecDEA}

\subsection{The framework of DEA-PPM}\label{framework}

\begin{algorithm}[!htb]
  \caption{The framework of DEA-PPM}\label{Alg_GCP}
  \KwIn{$G=(V,E)$, color number $k$}
  \KwOut{$\mathbf{x}^*_{G}$}

   initialize $\mathbf{Q}(0)$;   \\
   sample $\mathbf{Q}(0)$ to generate $\mathbf{P}(0)$; $/*$ \emph{uinform initialization} $*/$ \\

   denote the optimal solution in $\mathbf{P}(0)$ as $\mathbf{x}^*_{G}$\;
   %$\mathbf{q}^*_{G'}\leftarrow\mathbf{q}_1$, $\mathbf{x}^*_{G'}\leftarrow\mathbf{x}_1${\color{red}???????}\;
   $\mathbf{p}_1=\mathbf{x}^*_{G}$, $\mathbf{p}_2=\mathbf{x}^*_{G}$\;
   $t\leftarrow 1$\;
  %set $\mathbf{p}_1=\mathbf{x}^*_{G'}$, $\mathbf{p}_2=\mathbf{x}^*_{G'}$, $\mathbf{c}_1=\mathbf{x}^*_{G'}$\;

  \While {termination-condition 1 is not satisfied}
  {
        $\mathbf{Q}'(t)=OrthExpQ(\mathbf{Q}(t-1),\mathbf{P}(t-1))$; $/*$ \emph{orthogonal exploration} $*/$ \\
        $\mathbf{P}'(t)=SampleP(\mathbf{Q}'(t),\mathbf{P}(t-1)$; $/*$ \emph{ sampling with inheritance} $*/$\\
        $(\mathbf{P}(t),\mathbf{p}_1,\mathbf{p}_2,\mathbf{x}^*_{G})=RefineP(\mathbf{P}'(t),\mathbf{p}_1,\mathbf{p}_2,\mathbf{x}^*_{G})$\; $/*$ \emph{ refinement of the solution population} $*/$\\	
        %update $\mathbf{x}^{*}_{G'}$ and $\mathbf{q}^{*}_{G'}$ by $\mathbf{P}(t)$ and $\mathbf{Q}'(t)$, respectively\;
        	
        $\mathbf{Q}(t)=RefineQ(\mathbf{P}'(t),\mathbf{P}(t),\mathbf{Q}'(t))$; $/*$ \emph{ refinement of the distribution population} $*/$\\
        $t\leftarrow t+1$\;	
  }

\end{algorithm}

The framework of DEA-PPM for GCPs is presented in Algorithm \ref{Alg_GCP}~\cite{xu2022distribution}. It is implemented based on a distribution population $\mathbf{Q}(t)=(\mathbf{q}^{[1]}(t),\dots, \mathbf{q}^{[np]}(t))$ and a solution population $\mathbf{P}(t)=(\mathbf{x}^{[1]}(t),\dots, \mathbf{x}^{[np]}(t))$, where $\mathbf{x}^{[i]}(t)$ corresponds to $\mathbf{q}^{[i]}(t)$ one by one for all $i\in\{1,\dots,np\}$. For $k$-colroing of a decompostion graph $G$, DEA-PPM first initializes the distribution population $\mathbf{Q}(0)$ and the solution population $\mathbf{P}(0)$, and color $G$ by the iterative loop illustrated by Lines 6-12 of Algorithm \ref{Alg_GCP}.

The iterative loop attempts to obtain the optimal $k$-coloring assignment of $G$ by simultaneously evolving the distribution population $\mathbf{Q}(t)$ and the solution population $\mathbf{P}(t)$. In order to obtain the enhanced global exploration, it first performs orthogonal exploration on $\mathbf{Q}(t)$ to obtain $\mathbf{Q}'(t)$. Then, it generates an intermediate solution population $\mathbf{P}'(t)$ and performs a refinement process on $\mathbf{P}'(t)$ to obtain $\mathbf{P}(t + 1)$.

Additionally, $\mathbf{Q}'(t)$ is refined to obtain $\mathbf{Q}(t + 1)$, and the contents of the loop body are repeatedly executed, updating $\mathbf{x}^*_{G}$ until termination condition 1 is met. While numbers of conflicts and stitches are optimized to zero or the maximum number of iterations is reached, the iteration process is ceased and it outputs the corresponding $k$-coloring assignment of $G$. Details of the DEA-PPM is referred to Ref. \cite{xu2022distribution}.

%{\color{red} to be continued}
\subsection{Refinement of the solution population}\label{ls}
%DEA-PPM further refines the solution population by  $RefineP(\mathbf{P}'(t),\mathbf{p}_1,\mathbf{p}_2,\mathbf{x}^*_{G})$.
Because the MPLD problem minimizes the number of stitches as well as that of the conflicts, model (\ref{3}) of MPLD is slightly different with that of the GCP. Accordingly, the refinement process of DEA-PPM, presented in Algorithm \ref{Alg_Ref} is slightly modified to accommodate the MPLD problem well.

\begin{algorithm}[!htb]
\caption{$RefineP(\mathbf{P},\mathbf{p}_1,\mathbf{p}_2,\mathbf{x}^*_{G})$}\label{Alg_Ref}

      \KwIn{$\mathbf{P},\mathbf{p}_1,\mathbf{p}_2,\mathbf{x}^*_{G}$}
      \KwOut{$\mathbf{P},\mathbf{p}_1,\mathbf{p}_2,\mathbf{x}^*_{G}$}
      $iter\leftarrow 0$, $iter\_stag\leftarrow 0$\;
      $\mathbf{c}_1=\mathbf{x}^*_{G'}$\;
      $w_{2}\leftarrow 0$

      \While {$iter\_stag<6$}
    {
      $\mathbf{P}'=MGPX(\mathbf{P},\mathbf{p}_1,\mathbf{p}_2)$\;
      $\mathbf{P}'=Tabu(\mathbf{P}')$\;
      record the best solution in $\mathbf{P}'$ as $\mathbf{b}$\;
      \eIf{$f(\mathbf{b})<f(\mathbf{p}_1)$}
      { $w_{2} = 1$\;
      $iter\_stag=0$\;
      $\mathbf{c}_1=\mathbf{p}_1,\mathbf{p}_1=\mathbf{b}$\;}
      {$iter\_stag=iter\_stag+1$\;}

      \If{$f(\mathbf{b})<f(\mathbf{x}^*_{G'})$}
      {$\mathbf{x}^*_{G'}=\mathbf{b}$\;
      }

      \If{$mod(iter,3)==0 \&\& w_{2} == 1$}
      {$\mathbf{p}_2=\mathbf{c}_1$\;
      $w_{2} = 0$\;
      }
      %$(\mathbf{p}_1,\mathbf{p}_2,without)=Updatep(\mathbf{p}_1,\mathbf{p}_2,without,\mathbf{b},iter)$\;
      $\mathbf{P}=\mathbf{P}'$\;
      $iter=iter+1$\;
      }
      $\mathbf{P}'=Tabu(\mathbf{P})$\;
      $\mathbf{p}_1=\mathbf{b}$\;
\end{algorithm}

As presented in Algorithm \ref{Alg_Ref}, the iteration budget of \emph{while} loop is set as $6$ in this research.  Each iteration of the \emph{while} loop starts with a multi-parent greedy partition crossover (MGPX) and a tailored tabu search (TS), and ends with update of the solutions $\mathbf{p}_1$, $\mathbf{p}_2$ and $\mathbf{x}^*_{G}$. The MGPX is referred to Ref.~\cite{xu2022distribution}, and the TS process is presented in Algorithm \ref{Alg_ts}. The TS process for all $\mathbf{x}\in\mathbf{P}$ starts with initialization of the tabu list $\mathbf{T}$, which is a $k\times n$ matrix where element $T_{jv}$ represents the tabu level of coloring vertex $v$ with color $j$. Then, an iterative process is performed to  improve the quality of $\mathbf{x}$ and to update the tabu list $\mathbf{T}$.

\begin{algorithm}[ht]
      \caption{$Tabu(\mathbf{P})$}\label{Alg_ts}
      \KwIn{$\mathbf{P}$}
      \KwOut{$\mathbf{P}$}
      \For {$\mathbf{x}\in \mathbf{P}$}
    {
     $count\leftarrow 0$\;
     $\mathbf{y}=\mathbf{x}$\;
     initialize the tabu list $\mathbf{T}$ to a $k\times n$ zero matrix\;
     \While {termination-condition 2 is not satisfied}
      {
            choose the best legal mutation $<v,i,j>$\;
            perform the $<v,i,j>$ in $\mathbf{y}$\;
            Update $T_{jv}$ of the tabu list $\mathbf{T}$\;
             \If{$f(\mathbf{y})<f(\mathbf{x})$}
             {$\mathbf{x}=\mathbf{y}$\;}
            $count=count+1$\;
      }
     }
\end{algorithm}

Improvement of a solution $\mathbf{x}$ is achieved by the \emph{promising} mutations that generate promising solutions. For a vertex $v$ assigned with color $i$, a random mutation $<v,i,j>$ ($j\neq i$) is performed to assign $v$ with color $j $. While $<v,i,j>$ is not forbidden by the tabu list $\mathbf{T}$, it generates a candidate solution that would replace the coloring scheme $\mathbf{x}$. To get solutions improved as far as possible, the best promising mutation that contributes to the smallest numbers of conflicts and stitches is employed to generate the promising solution $\mathbf{y}$, by which the present solution $\mathbf{x}$ is updated if $f(\mathbf{y})<f(\mathbf{x})$.

 A random mutation $<v,i,j>$ is forbidden by the tabu list if the iteration number $count$ is less than $T_{jv}$. If a  solution $\mathbf{y}$ is generated by implementing  $<v,i,j>$, the corresponding element of $T$  is updated by
  \begin{equation*}
  T_{j,v}=count+0.6*(10*y_{c}+y_{s})+r,
\end{equation*}
where $count$ is the iteration number of TS, $y_{c}$ is the number of conflicts in $\mathbf{y}$, $y_{s}$ is the number of stitches in $\mathbf{y}$, and $r$ is an integer randomly sampled in $\{1,\dots,10\}$.

The iteration of TS is repeated until  \emph{termination condition 2} is satisfied, that is,  a solution without stitches and conflicts is obtained or the number of iterations reaches the iteration budget $5*|V|$.

\section{Experimental Results}\label{SecRes}
\begin{figure*}[!hbt]
\centering
\subfigure[The result of TPLD.]{
\includegraphics[width=7.5cm]{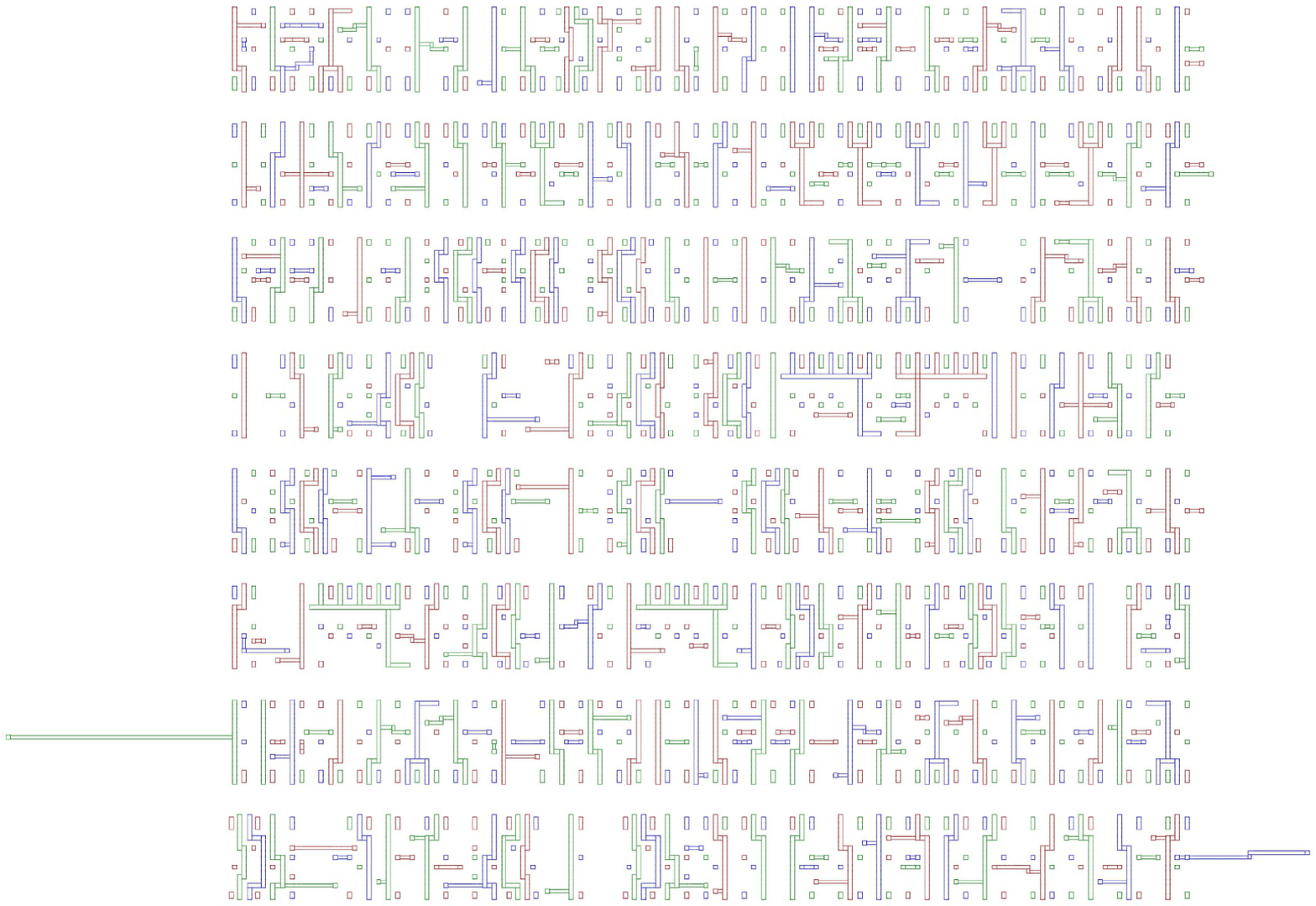}
}\label{p1}
%\quad
\subfigure[The result of QPLD.]{
\includegraphics[width=7.5cm]{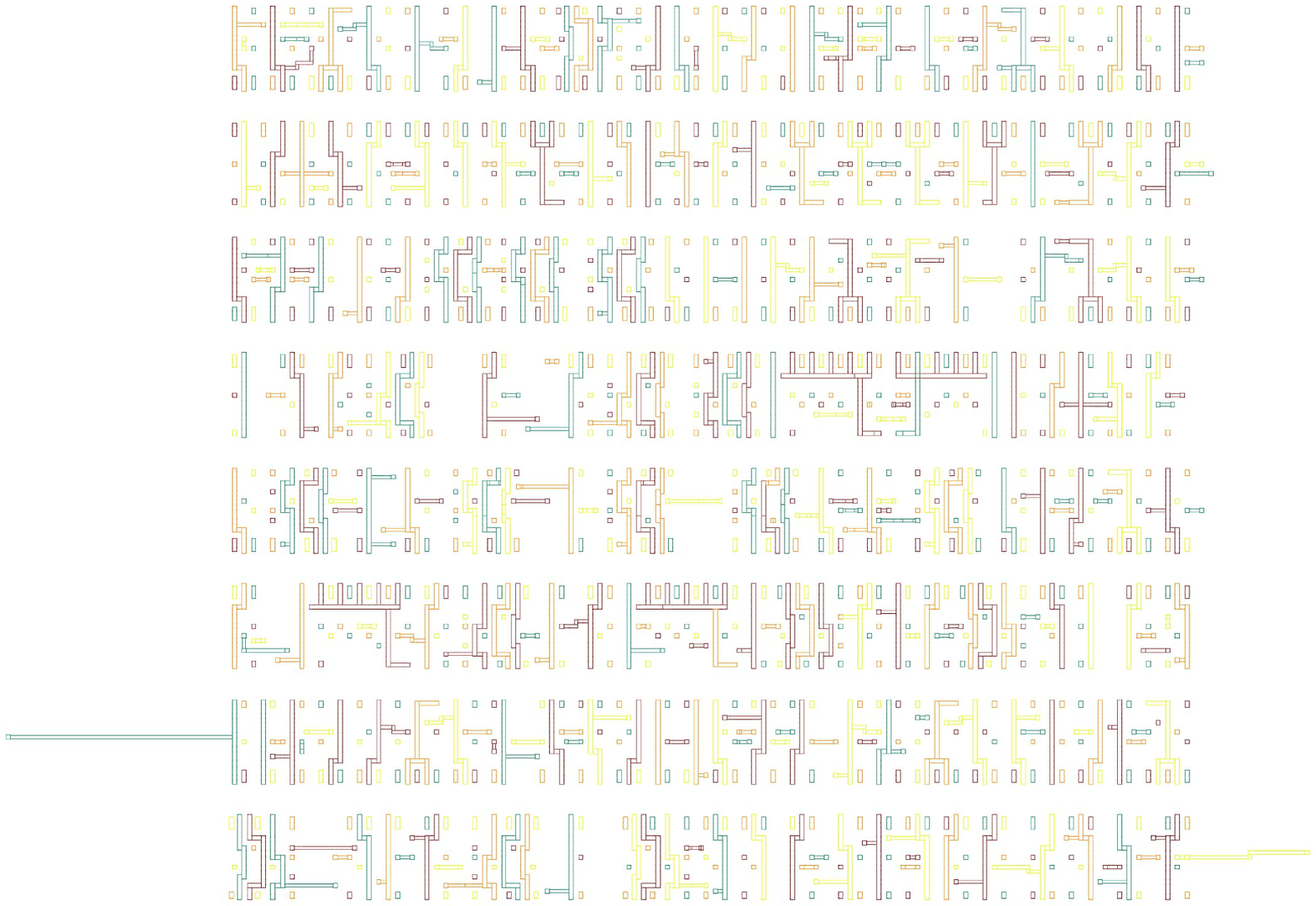}
}\label{p2}
\caption{Decomposition results of the layout benchmark \emph{C432} obtained by the DEA-PPM.}
\label{decomposition}
\end{figure*}

%\begin{figure*}[!hbt]
%\centering
%\subfigure[Decomposed layout for TPLD with $\min_{cs}$=100nm.]{
%\includegraphics[width=8cm]{p1.eps}
%}\label{p1}
%%\quad
%\subfigure[Decomposed layout for QPLD with $\min_{cs}$=200nm.]{
%\includegraphics[width=8cm]{p2.eps}
%}\label{p2}
%\caption{Decomposed layouts of the benchmark \emph{C432} obtained by the DEA-PPM.}
%\label{decomposition}
%\end{figure*}
Competitiveness of the proposed MPLD algorithm is validated by result comparison for the ISCAS benchmark problems. The \emph{OpenMPL} decomposer is implemented by C++ on a laptop equipped with Intel Core 1.10 GHz CPU and a virtual Linux machine. By testing the performance of DEA-PPM on the ISCAS benchmarks, we compare the results by the stitch numbers(`st\#'), the conflict numbers(`cn\#'), the cost values(`cost') and the CPU running times(`time') of ILP, SDP and DEA-PPM. Because DEA-PPM is a stochastic optimization algorithm, the reported results are average values of 10 independent runs, and we also investigate the standard deviations of results to demonstrate its robustness. The scalability of DEA-PPM is validated by varied settings of both the number of masks and the minimum coloring space $min_{cs}$. To perform a fair comparison, the \emph{OpenMPL} decomposer is run in the single-thread mode. An illustration of the decomposition results is presented for DEA-PPM in Fig. \ref{decomposition}.

\subsection{Comparison for the TPLD problem with $min_{cs}=120/100$nm }\label{FIR}

As presented in \cite{9279240}, \emph{OpenMPL} works well for TPLD of the ISCAS benchmarks, where the minimum coloring spacing $min_{cs}$ is set as \emph{120nm} for the benchmarks \emph{C432-C7552} and \emph{100nm} for the benchmarks \emph{S1488-S15850}. Then, we compare the decomposition results of DEA-PPM with those of the ILP and the SDP.

The decomposition results of three algorithms are presented in Tab. \ref{1}. For the small value of $min_{cs}$, vertexes of the decomposition graph would be incident to less edges. The results demonstrate that the proposed DEA-PPM, as a stochastic optimization algorithm, can always achieve the same results as ILP with less running time. Because the SDP method addresses a continuously relaxed problem of model (\ref{model}), it runs faster than ILP and DEA-PPM, but cannot always converge to the global optimal solutions at all time. Consequently, the decomposition results obtained by the SDP method are generally the same as or worse than those of ILP and DEA-PPM.

\begin{figure}[!hbt]
\centering
\includegraphics[width=8cm]{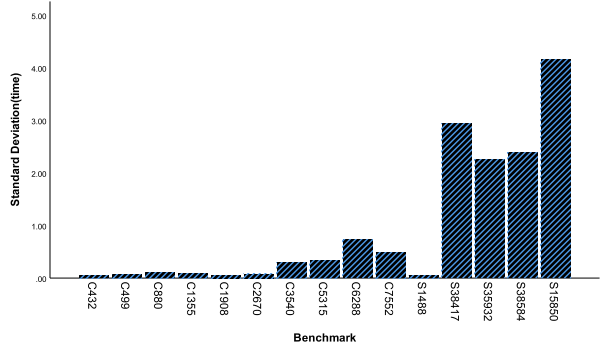}
\caption{Standard deviations of running time for the TPLD problem with $min_{cs}=120/100 nm$.}
\label{std_dev}
\end{figure}

It is surprising that the standard deviation of cost is zero, which shows that all independent runs of DEA-PPM converge to the  global optimal solution of model (\ref{model}). Furthermore, the standard deviations of running time are collected in Fig. \ref{std_dev}. It is demonstrated that the standard deviation of running time is ignorable for most of the benchmark layouts, except that the standard deviations of running time are about a few seconds for the cases \emph{S35392}, \emph{S38584} and \emph{S15850}.

\begin{table}[!htp]
\centering
\caption{Result comparison for the TPLD with $min_{cs}=120/100nm$}
\resizebox{\hsize}{!}{
\begin{tabular}{l|cccc|cccc|cccc}
\hline \hline
\multirow{2}{*}{Circuit} & \multicolumn{4}{c|}{ILP}        & \multicolumn{4}{c|}{SDP}        & \multicolumn{4}{c}{DEA-PPM}
                         \\
                         \cline{2-13}
        & st\# & cn\# & cost & time(s) & st\# & cn\# & cost & time(s) & st\# & cn\# & cost  & time(s)  \\
\hline
C432   & 4   & 0  & 0.4  & 0.38  & 4   & 0  & 0.4  & 0.09  & 4   & 0  & 0.4  & 0.30  \\
C499   & 0   & 0  & 0    & 0.31  & 0   & 0  & 0    & 0.13  & 0   & 0  & 0    & 0.25  \\
C880   & 7   & 0  & 0.7  & 0.52  & 7   & 0  & 0.7  & 0.14  & 7   & 0  & 0.7  & 0.46  \\
C1355  & 3   & 0  & 0.3  & 1.16  & 3   & 0  & 0.3  & 0.17  & 3   & 0  & 0.3  & 0.59  \\
C1908  & 1   & 0  & 0.1  & 0.6   & 1   & 0  & 0.1  & 0.28  & 1   & 0  & 0.1  & 0.52  \\
C2670  & 6   & 0  & 0.6  & 1.29  & 6   & 0  & 0.6  & 0.51  & 6   & 0  & 0.6  & 0.60  \\
C3540  & 8   & 1  & 1.8  & 2.31  & 8   & 1  & 1.8  & 0.82  & 8   & 1  & 1.8  & 1.59  \\
C5315  & 9   & 0  & 0.9  & 2.02  & 9   & 0  & 0.9  & 1.49  & 9   & 0  & 0.9  & 1.84  \\
C6288  & 205 & 1  & 21.5 & 12.57 & 203 & 7  & 27.3 & 4.32  & 205 & 1  & 21.5 & 10.48 \\
C7552  & 23  & 0  & 2.3  & 6.97  & 23  & 0  & 2.3  & 1.38  & 23  & 0  & 2.3  & 3.27  \\
S1488  & 2   & 0  & 0.2  & 0.58  & 2   & 0  & 0.2  & 0.31  & 2   & 0  & 0.2  & 0.42  \\
S38417 & 54  & 19 & 24.4 & 17.62 & 46  & 27 & 31.6 & 9.31  & 54  & 19 & 24.4 & 16.13 \\
S35932 & 40  & 44 & 48   & 43.49 & 20  & 64 & 66   & 27.5  & 40  & 44 & 48   & 32.18 \\
S38584 & 116 & 36 & 47.6 & 41.37 & 105 & 48 & 58.5 & 28.37 & 116 & 36 & 47.6 & 35.83 \\
S15850 & 97  & 34 & 43.7 & 35.53 & 83  & 48 & 56.3 & 22.62 & 97  & 34 & 43.7 & 33.68 \\
\hline \hline
\end{tabular}
}
\label{1}
\end{table}

\subsection{Comparison for the TPLD problem with $min_{cs}=160$nm}\label{SEC}
To demonstrate the scalability of our proposed method for various lithography technologies, we investigate the TPLD of ISCAS benchmarks by setting the minimum coloring spacing $min_{cs}=160 nm$. Because the lithography resolution is lower, there could be more conflicting edges in the layout graph, and more stitches would be inserted to get the decomposition graph colorable. Accordingly, the community structure of the decomposition graph could be fuzzier, and decomposition of the layouts is more challenging.

\begin{figure}[!hbt]
\centering
\subfigure[Standard deviations of cost values.]{
\includegraphics[width=7.5cm]{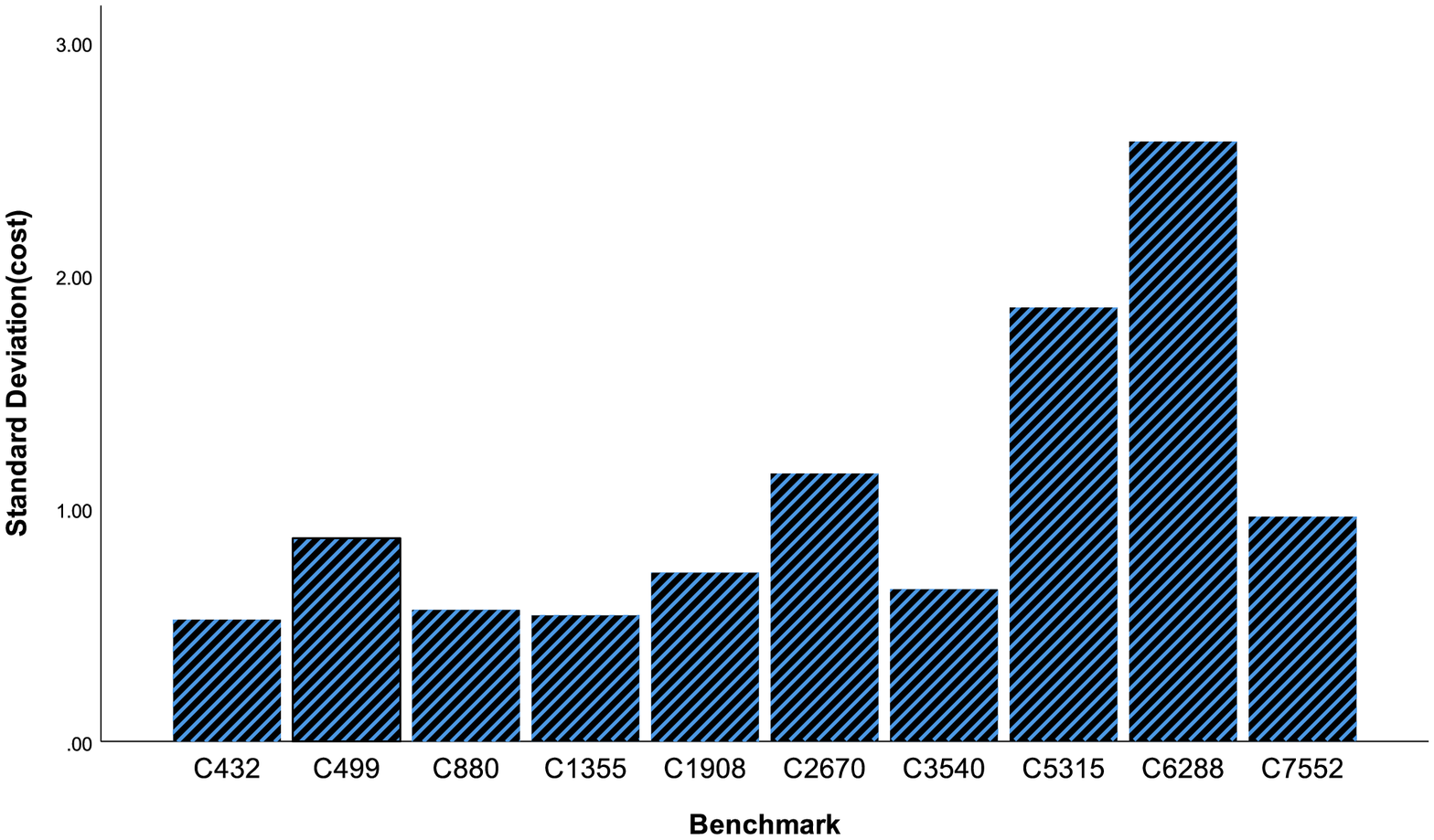}
}\label{std2}
\subfigure[Standard deviations of running time.]{
\includegraphics[width=7.5cm]{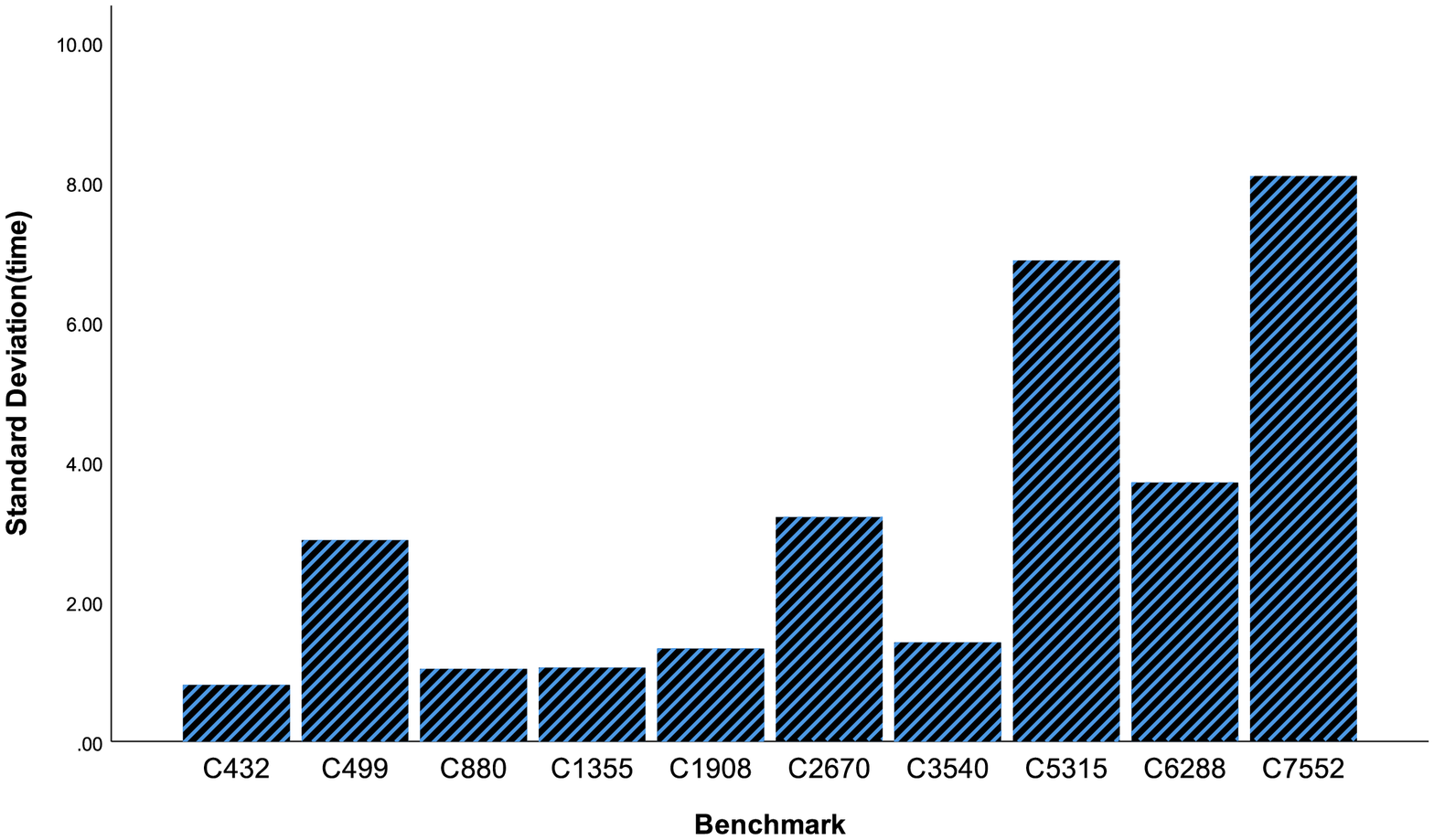}
}\label{std3}
\caption{Standard deviations of results obtained by the DEA-PPM for the TPLD problem with $min_{cs}=160 nm$.}
\label{std_dev1}
\end{figure}

The experimental results collected in Table \ref{2} show that DEA-PPM can deal well with the low-resolution TPLD problem. Compared with the ILP, DEA-PPM contributes to a decrease of more than one order of magnitude of the running time at the expense of a bit increase of the cost values. Meanwhile, the cost values optimized by DEA-PPM are much smaller than the results of SDP, while the running time of DEA-PPM is a bit greater that of SDP.  The histograms illustrated in Fig. \ref{std_dev1} show that the standard deviations of cost values are smaller than $3$, and those of running time are not more than $10$ seconds. Accordingly, DEA-PPM is also robust when applied to address the TPLD problem with $min_{cs}=160 nm$.

\begin{table}[!htp]
\centering
\caption{Result comparison for the TPLD with $min_{cs}=160nm$}
\resizebox{\hsize}{!}{
\begin{tabular}{l|cccc|cccc|cccc}
\hline \hline
\multirow{2}{*}{Circuit} & \multicolumn{4}{c|}{ILP}         & \multicolumn{4}{c|}{SDP}          & \multicolumn{4}{c}{DEA-PPM}                                \\
\cline{2-13}
        & st\# & cn\# & cost  & time(s) & st\# & cn\# & cost   & time(s) & st\#  & cn\#  & cost    & time(s)  \\
\hline
C432  & 17  & 76  & 77.7  & 113.02  & 19  & 80  & 81.9   & 5.29  & 17.1  & 76.9  & 78.61  & 6.92  \\
C499  & 48  & 278 & 282.8 & 339.77  & 48  & 279 & 283.8  & 10.48 & 48    & 278.2 & 283    & 16.40 \\
C880  & 123 & 103 & 115.3 & 244.9   & 113 & 115 & 126.3  & 9.26  & 122.4 & 103.7 & 115.94 & 12.06 \\
C1355 & 116 & 114 & 125.6 & 356.94  & 109 & 122 & 132.9  & 21.59 & 118.3 & 113.9 & 125.73 & 17.26 \\
C1908 & 109 & 150 & 160.9 & 136.82  & 100 & 162 & 172    & 12.18 & 106.2 & 150.7 & 161.32 & 16.95 \\
C2670 & 369 & 354 & 390.9 & 3506.45 & 358 & 386 & 421.8  & 35.71 & 369.4 & 357.2 & 394.14 & 41.05 \\
C3540 & 506 & 334 & 384.6 & 368.65  & 482 & 362 & 410.2  & 25.08 & 504   & 336.4 & 386.8  & 40.27 \\
C5315 & 487 & 779 & 827.7 & 975.58  & 467 & 819 & 865.7  & 42.6  & 487   & 786.2 & 834.9  & 71.04 \\
C6288 & 384 & 616 & 654.4 & 878.71  & 366 & 653 & 689.6  & 53.76 & 386.1 & 622.3 & 660.91 & 58.47 \\
C7552 & 858 & 843 & 928.8 & 3376.73 & 808 & 933 & 1013.8 & 68.12 & 851.1 & 850.6 & 935.71 & 89.28 \\
\hline \hline
\end{tabular}
}
\label{2}
\end{table}

\subsection{Comparison for the QPLD problem with $min_{cs}=200 nm$}\label{THI}

\begin{figure}[!hbt]
\centering
\subfigure[Standard deviations of cost values.]{
\includegraphics[width=7.5cm]{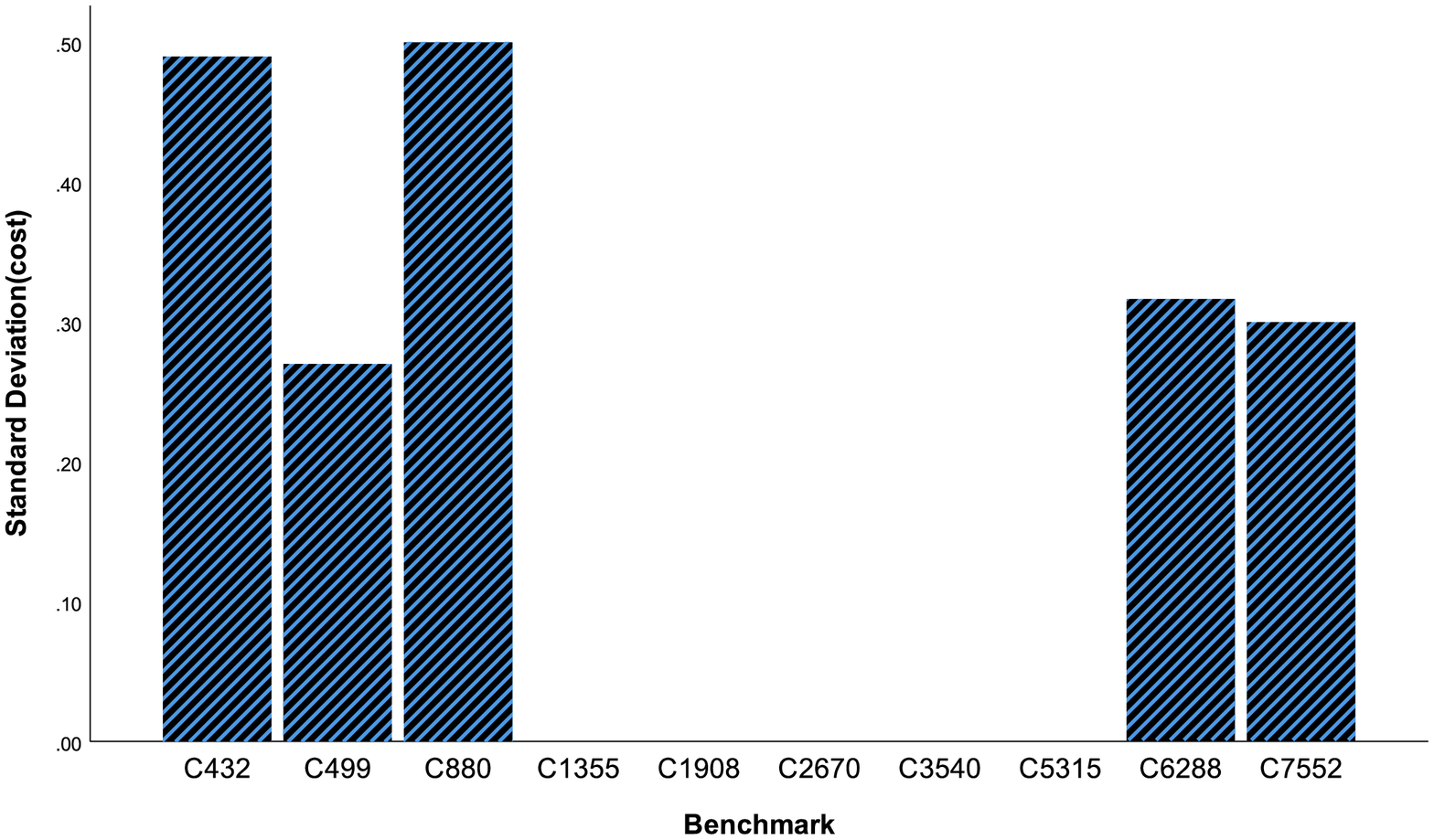}
}\label{std4}
\subfigure[Standard deviations of running time.]{
\includegraphics[width=7.5cm]{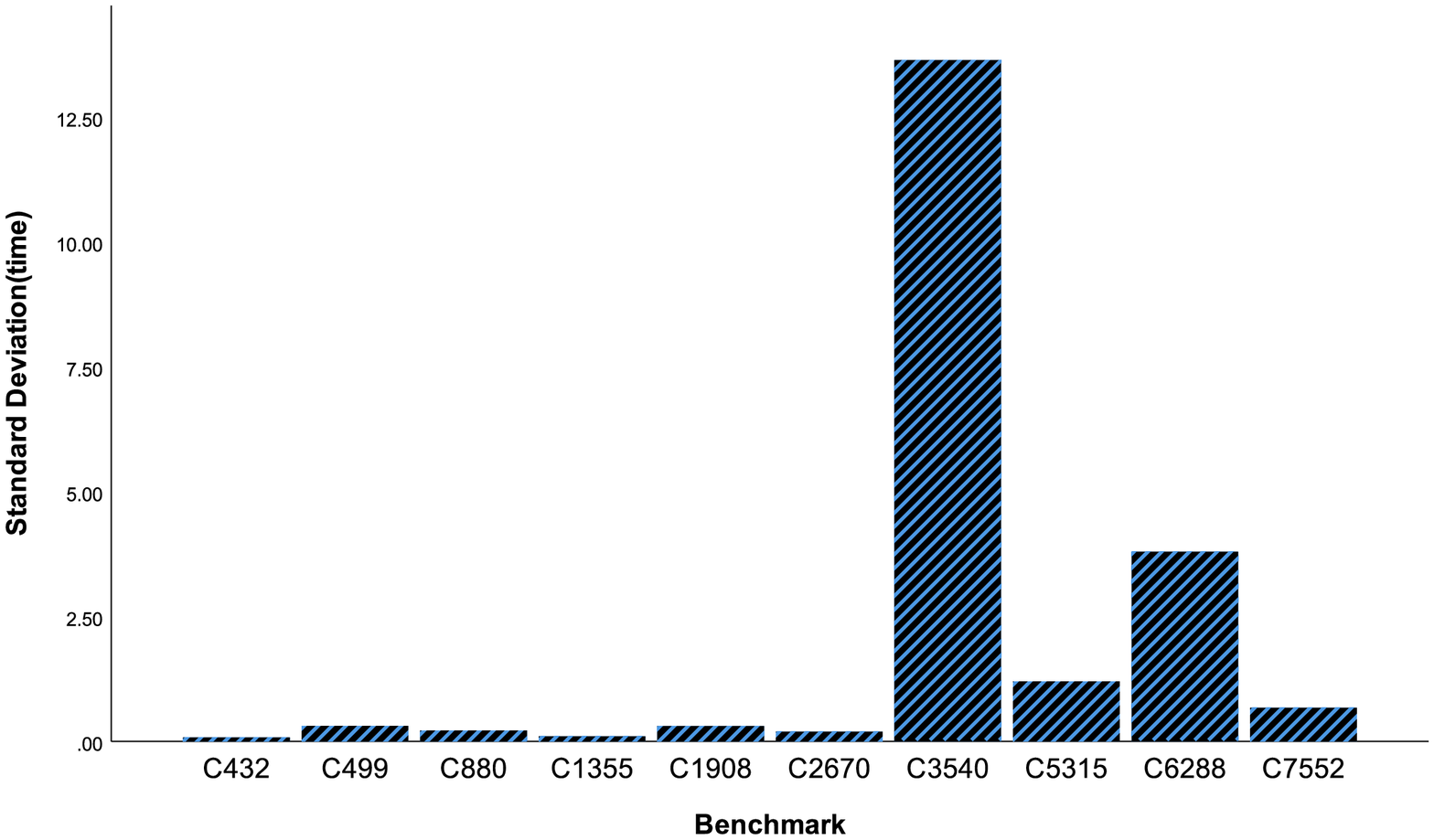}
}\label{std5}
\caption{Standard deviations of results obtained by the DEA-PPM for the QPLD problem with $min_{cs}=200 nm$.}
\label{std_dev2}
\end{figure}

Besides varied settings of the lithography resolution, we further test the scalability of DEA-PPM  by investigating the QPLD problem, where the minimum coloring spacing is set as $200nm$ for 10 ISCAS benchmarks. The test results are listed in Table \ref{3}, and Fig. \ref{std_dev2} illustrates the standard deviations of both cost values and running time for 10 independent runs. Since the ILP cannot address the \emph{C6288} benchmark in 1 hour, we denote the corresponding results by
‘-’.

The obtained cost values of DEA-PPM are generally better than that of SDP, and its running time is much shorter than that of ILP.  It is surprising that compared with the ILP, DEA-PPM achieves better cost values on cases \emph{C432}, \emph{C6288} and \emph{C7552} with much shorter running time. Although the running time of DEA-PPM is a bit longer than that of SDP, the obtained better cost values still demonstrate its competitiveness on the QPLD problem. Fig. \ref{std_dev2} demonstrates that the standard deviations of cost value is about $0.5$ for all $10$ benchmarks, and the running time varies with standard deviations less than $5$ seconds for most benchmarks except for \emph{C3540}.

\begin{table}[!htp]
\centering
\caption{Result comparison for the QPLD with $min_{cs}=200nm$}
\resizebox{\hsize}{!}{
\begin{tabular}{l|cccc|cccc|cccc}
\hline \hline
\multirow{2}{*}{Circuit} & \multicolumn{4}{c|}{ILP}        & \multicolumn{4}{c|}{SDP}        & \multicolumn{4}{c}{DEA-PPM}                             \\
\cline{2-13}
        & st\# & cn\# & cost & time(s) & st\# & cn\# & cost & time(s) & st\# & cn\# & cost   & time(s) \\
\hline
C432  & 4  & 14 & 14.4 & 61.38  & 4  & 12  & 12.4 & 1.61  & 4    & 13.6  & 14     & 1.79  \\
C499  & 29 & 7  & 9.9  & 123.71 & 29 & 8   & 10.9 & 8.14  & 28.9 & 7.1   & 9.99   & 4.72  \\
C880  & 4  & 6  & 6.4  & 4.1    & 4  & 7   & 7.4  & 0.58  & 4    & 6.5   & 6.9    & 1.36  \\
C1355 & 4  & 12 & 12.4 & 13.44  & 4  & 12  & 12.4 & 1.41  & 4    & 12    & 12.4   & 1.65  \\
C1908 & 10 & 24 & 25   & 50.42  & 10 & 24  & 25   & 3.94  & 10   & 24    & 25     & 3.98  \\
C2670 & 11 & 24 & 25.1 & 225.96 & 12 & 25  & 26.2 & 5.98  & 11   & 24    & 25.1   & 4.25  \\
C3540 & 18 & 39 & 40.8 & 49.94  & 20 & 43  & 45   & 4.26  & 18   & 39    & 40.8   & 46.22  \\
C5315 & 31 & 38 & 41.1 & 90.47  & 30 & 40  & 43   & 5.63  & 31   & 38    & 41.1   & 8.77  \\
C6288 & -  & -  & -   & -      & 10 & 299 & 300  & 18.32 & 8.3  & 297.9 & 298.73 & 31.09 \\
C7552 & 38 & 60 & 63.8 & 706.97 & 36 & 61  & 64.6 & 7.66  & 38   & 59.1  & 62.9   & 11.48 \\
\hline \hline
\end{tabular}
}
\label{3}
\end{table}

\section{Conclusion}\label{SecCon}

To develop a scalable layout decomposition scheme adapted for a variety of lithography resolution with different number of masks, we proposed to model the huge-scale decomposition problem as an extended graph coloring problem, which is addressed by a distribution evolutionary algorithm based on a population of probability models (DEA-PPM).

Since the DEA-PPM employs a distribution evolution scheme assisted by the tabu search, it is scalable to varied settings of the MPL layout decomposition problem. Experimental results on ISCAS benchmarks demonstrate that DEA-PPM can strike a good balance between the optimization results and the running time. Despite that DEA-PPM is generally superior to ILP and inferior to SDP in terms of running time, it can obtain for some benchmarks both smaller cost values and less running time, which show that the proposed decomposition method is competitive to the existing decomposition schemes.

Despite the stochastic nature of DEA-PPM, its tailored search strategies leads to small standard deviations of decomposition results, which shows that the proposed method could be an alternative method for the MPLD of layout.

\section*{Acknowlegement}
We thank Dr. Ning Li  for his constructive suggestions to this work.



\bibliographystyle{elsarticle-num}
\bibliography{mybibfile}


%\begin{thebibliography}{00}


%\bibitem{}

%\end{thebibliography}
\end{document}